\documentclass{llncs}

\newcommand{\paperTitle}{Automatically Discovering Hidden Transformation Chaining Constraints}

\usepackage{graphicx}
\usepackage[utf8]{inputenc}
\usepackage[T1]{fontenc}
\usepackage[bookmarks=false,breaklinks=true,pdfstartview=FitH,pdftitle={\paperTitle},pdfauthor={Rapha\"el Chenouard, and Fr\'ed\'eric Jouault}]{hyperref}
\usepackage{color}
\usepackage{listings}
\usepackage{rotating}

\usepackage{AMMALanguages}

\newcommand{\sCOMMA} {\textsf{\footnotesize{s-COMMA}}}
\newcommand{\Eclipse} {ECL$^{i}$PS$^{e}$}

\title{\paperTitle}
\author{Rapha\"el Chenouard\inst{1} \and Fr\'ed\'eric Jouault\inst{2}}
\institute{LINA, CNRS, Universit\'e de Nantes, France\\
\and
AtlanMod (INRIA \& EMN), France\\
{\texttt{raphael.chenouard@univ-nantes.fr}},
\texttt{frederic.jouault@inria.fr}}

\begin{document}
\maketitle

\begin{abstract}
Model transformations operate on models conforming to precisely defined
metamodels. Consequently, it often seems relatively easy to chain them: the
output of a transformation may be given as input to a second one if metamodels
match. However, this simple rule has some obvious limitations. For instance, a
transformation may only use a subset of a metamodel. Therefore, chaining
transformations appropriately requires more information.

We present here an approach that automatically discovers more detailed
information about actual chaining constraints by statically analyzing
transformations. The objective is to provide developers who decide to chain
transformations with more data on which to base their choices. This approach
has been successfully applied to the case of a library of endogenous
transformations. They all have the same source and target metamodel but have
some hidden chaining constraints. In such a case, the simple metamodel matching
rule given above does not provide any useful information.
\end{abstract}

\section{Introduction}

One of the main objectives of Model-Driven Engineering (MDE) is to automatize
software engineering tasks such as:
the production of code from abstract models in forward engineering scenarios,
the production of  abstract models from code in reverse engineering scenarios,
or a combination of the two previous cases in modernization scenarios.
To achieve this automation, MDE relies on precisely defined models that can be
processed by a computer.
Each model conforms to a metamodel that defines concepts as well as relations
between them. For instance, a Java metamodel has the concept of Java class,
with the corresponding single-valued superclass relation (i.e., a class can only
extend one other class). Similarly, the UML metamodel defines the concept of a
UML class, with a multi-valued generalization relation (i.e., a class may extend
several other classes). Many software engineering tasks such as those mentioned
above can be performed by model transformations.

In order to reduce the effort of writing these transformations, complex tasks
are generally not performed by complex transformations but rather by chains of
simpler transformations. A model transformation chain is formed by feeding the
output of a first transformation as input to second one. Complex chains can
consist of a large number of transformations. For instance, in order to analyze
a Java model with a Petri net tool, a first transformation may operate from Java
to UML, and a second one from UML to Petri net.

Model transformations are reusable. 
In our previous example, if a different target formalism is to be used, the Java
to UML transformation may be reused, while the second one is replaced. Some
model transformation libraries such as \cite{ATLLibrary} are already available to leverage this
possibility. Typically, each entry specifies the name of the transformation as
well as its source and target metamodels. Some documentation may also be
available. A model-driven engineer confronted to a model transformation problem
may first lookup for existing transformations. If no pre-existing transformation
exactly performs the required task, some pieces may be used to form a chain into
which only simpler new transformations need to be inserted.
Source and target metamodel information may be used to chain transformations.
A transformation from B to C may for instance be attached at the output of a
transformation from A to B.

However, chaining transformations properly is generally a more complex task in
practice. Knowing the source and target metamodels of a transformation is not
enough. For instance, a transformation may only target a subset of its declared
target metamodel.
Feeding its output to a second transformation that takes a different subset of
the same metamodel as input will typically not yield correct results. Computing
a class dependency graph from a Java model by reusing a transformation that
takes UML Class diagrams as input may not be possible with the Java to UML
transformation targeting Petri nets used in the previous example. While this new
transformation requires the class structure to be retrieved from the Java model,
the initial transformation may have been limited to the generation of the
Activity diagrams required for the generation of Petri nets.

The case of endogenous transformations is even more problematic. Because these
transformations have the same source and target metamodel, they can in theory be
inserted in a chain anywhere this metamodel appears. A collection of such
transformations operating on the same metamodel could also be chained in any
order. In practice, this may not lead to correct results (e.g., because a transformation may remove an element from the model that is required for another transformation to perform correctly).

Chaining transformations actually requires more precise knowledge about the
individual transformations. For instance, if transformation $t_1$ relies on some information that is dropped by transformation $t_2$ then $t_1$ cannot be applied after $t_2$. This knowledge may be available in a documentation
of some sort, but this is not always the case. One may also look at the insides
of a transformation (i.e., its implementation), but this requires knowledge of
the transformation language (there are several languages, and not everybody is
an expert in all of them).

The situation would be simplified if each transformation clearly identified the
subset of a metamodel it considers. But this is not always enough. For instance,
some endogenous transformations have a fixed point execution semantics (i.e.,
they need to be executed again and again until the resulting model is not
changed any more). In such a case, the metamodel subset generated by each iteration may be different (especially the last iteration when compared to the previous ones for transformations that remove elements one at a time).

The purpose of the work presented here is to automatically discover information
about what model transformations actually do. The resulting data may be used to
help the engineer decide how to chain transformations, and may complement what
is in the documentation of the transformation if there is one.
A Higher-Order Transformation \cite{ECMDA2009} that takes as input the transformations to analyze produces a model containing the analysis results.
This model may then be rendered to various surfaces using other transformations.

We have applied this approach to the case of a set of endogenous transformations that are used for the translation between constraint programming languages.
All transformations take the same pivot metamodel as source and target metamodel and are written in ATL \cite{MTIP05b,SAC06a,SCP07c} (AtlanMod Transformation Language).
However, different subsets of this metamodel are actually consumed and produced by each transformation.
By statically analyzing these transformations we have been able to discover what they do, and infer chaining constraints from this knowledge.

The reminder of the paper is organized as follows.
Section \ref{sec:motivatingExample} presents a scenario involving a number of
endogenous transformations operating on a single metamodel.
Our transformation analysis approach is described in Section
\ref{sec:transformationAnalysis}, and its application is presented in Section
\ref{sec:experiments}.
The results are discussed in Section \ref{sec:discussions}.
Finally, Section \ref{sec:conclusion} concludes.

%===============================================================================
\section{Motivating Example}
\label{sec:motivatingExample}

\subsection{Interoperability of Constraint Programming Languages}
\begin{figure}[htpb]
\includegraphics[width=0.9\linewidth]{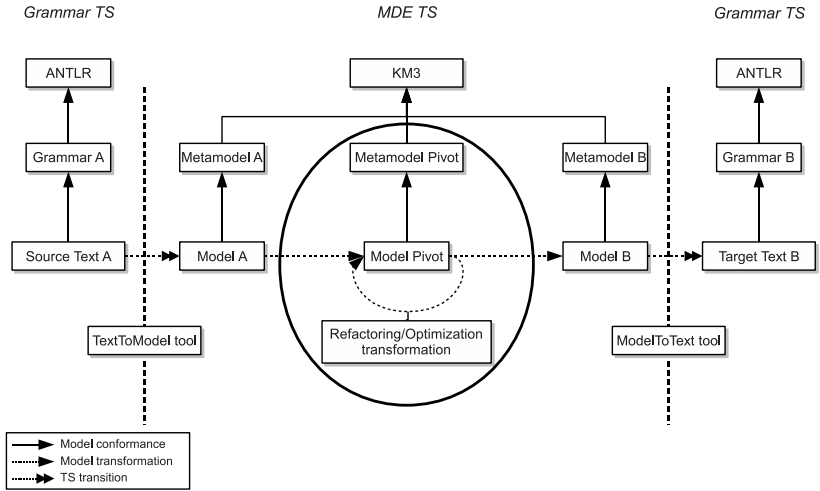}
\caption{A generic transformation process to translate CP
models.\label{fig:CP-Trans}}
\end{figure}
In Constraint Programming (CP), one of the main goals is to define problems based
on variables, domains and constraints such that a CP solver can compute their
solutions~\cite{RossiElsevier2006}. In CP, various kinds of languages are used
to state problems. For instance, the language of the \Eclipse{}
solver~\cite{EclipseBook2007} is based on logic and Prolog, whereas 
OPL~\cite{OPL1999} (Optimization Programming Language) is a solver-independent language based on high-level
modeling constructs. Some solvers have only programming APIs like ILOG
Solver~\cite{PugetSCIS1994} or Gecode~\cite{SchulteGecode2006}. More recently,
the definition of high-level modeling languages is becoming a hot topic in
CP~\cite{PugetCP04}. Then, new modeling languages have been developed
such as Zinc and MiniZinc~\cite{Zinc2008}, Essence~\cite{FrischIJCAI2007} and
\sCOMMA{}~\cite{SotoICTAI2007}. In these three cases, the high-level modeling
language is translated into existing CP solver languages by using a flat
intermediary language to ease the translation process and to increase the
reusability of most transformations and reformulation tasks. This process is
mainly achieved by hand-written translators using parsers and lexers.

In a recent work~\cite{ChenouardPPDP2008}, model engineering was used to carry
out this process from \sCOMMA{} models to some solver languages. Then, this
approach has been extended to get more freedom in the choice of the user
modeling language~\cite{ChenouardSara2009} (see Figure~\ref{fig:CP-Trans}). A
flexible pivot CP metamodel was introduced, on which several transformations are
performed to achieve generic and reusable reformulation or optimization steps.
The transformation chain from a language A to a language B is composed of three
main steps: from A to pivot, pivot refactoring and pivot to B. Steps on pivot
models may remove some structural features not authorized by the target solver
language. Thus, objects, if or loop statements may be removed and replaced by an
equivalent available structure, i.e. objects are flattened, if are expressed as
boolean expressions and loops are unrolled. All these refactoring steps are not
mandatory when considering a CP modeling language and a CP solver language,
since loop or if statements may be available in most CP solvers. Since no
existing model engineering tool exists to automate the chaining of these model
transformations according to a source and a target metamodel, the user must
build chains by hand without any verification process.

\begin{figure}[htpb]
\includegraphics[width=\linewidth]{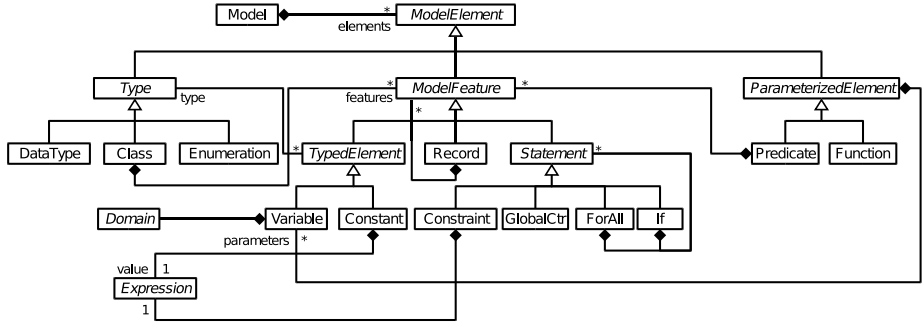}
\caption{Generic pivot metamodel for CP (excerpt).\label{fig:CP-Pivot}}
\end{figure}

The main part of the generic CP pivot metamodel introduced
in~\cite{ChenouardSara2009} is shown on Figure~\ref{fig:CP-Pivot}. Indeed, CP
models are composed of a set of constraints, variables and domains. They are
classified in an inheritance hierarchy, with abstract concepts such as
\texttt{Statement} that corresponds to all kinds of constraint declarations.
High-level model constructs are defined according to existing modeling
languages, such as the class and record concepts. Most of pivot models will
only contained elements conforming only to a subset of the whole pivot
metamodel.

\subsection{Problem}

In this paper, we want to tackle the issues relating to the efficient
management of a set of endogenous transformations. Since the source and
target metamodels are similar, no additional information can be
extracted from the header of an ATL transformation. Considering only
this knowledge, we may think that endogenous transformations can be
chained without any problem, but this is not true. The solution
proposed by~\cite{Vignaga2008} is therefore not sufficient to address this
problem because it only considers the signature (or header) of transformations. As shown in the motivating example, endogenous
transformations achieve model reformulation or optimization
steps. They have to be efficiently and correctly chained to avoid
useless steps --- some steps may create elements that are removed by
another step --- and to reach the requirements of the target solver
language. Our goal is to discover the role of endogenous model
transformations in a parameterizable chain.

Endogenous transformations can be typed using their source and target element
types, i.e. a sub-set of the metamodel of these models. Thus, considering the
set of source elements of an endogenous transformation, we can assess the set of
source models supported by it without any loss. The set of target elements also
allows us to type generated models. Then, we may be able to verify endogenous
transformation chains. Moreover, using a search/optimization algorithm we may be
able to find the "best" chain and thus automating the chaining of endogenous
transformations according to an input metamodel and to an output metamodel
corresponding to a high-level exogenous transformation.

%===============================================================================
\section{Transformation Analysis}
\label{sec:transformationAnalysis}

\subsection{Identifying Domains and Codomains}

In order to correctly chain model transformations it is necessary to have a certain understanding of what they do.
Although it is not enough, source and target metamodels information is essential.
The model $M_B$ produced by a given transformation $t_1$ conforms to its target metamodel $MM_B$.
It may only be fed as input to another transformation $t_2$ with the same metamodel $MM_B$ as source metamodel.

This constraint may be expressed in functional terms as shown in~\cite{Vignaga2008}: transformations are considered as functions, and metamodels type their parameters in the case of simple transformations (Higher).
For instance, if the source metamodel of $t_1$ is $MM_A$, and the target metamodel of $t_2$ is $MM_C$ then:
$t_1 : MM_A \rightarrow MM_B$ and $t_2 : MM_B \rightarrow MM_C$.
In this notation the name of a metamodel is used to identify the set of models that conform to it.
Thus, transformation $t_1$ is considered as a function of domain the set of models conforming to $MM_A$, and of codomain the set of models conforming to $MM_B$.
In this example, if $t_2$ is total then it may be applied to the output of $t_1$ because the codomain of $t_1$ is also the domain of $t_2$.

In practice, model transformations are often partial functions: they do not map every element of their declared domain to an element of their codomain.
For instance, $t_2$ may only work for a subset $MM_B' \subset MM_B$.
If $t_1$ is surjective (i.e., it can produce values over its whole codomain) then $t_2$ cannot be applied to all output models that $t_1$ can
 produce.
This shows that problems can arise when the domain of transformations (i.e., their source metamodels) is underspecified (i.e., too broad).
If codomains (or target metamodels) are also underspecified, then there may not be any actual problem.
For instance, if $t_1$ only produces results over $MM_B'' \subseteq MM_B'$ then $t_2$ may be chained to $t_1$.
Therefore, precisely identifying the actual domain and codomain of a transformation (i.e., definition domain and its image) would be an improvement over the current practice.

However, doing so is often complex because it requires deep analysis of transformations (e.g., not only source elements of transformation rules but also every navigation over source elements).
Moreover, the semantics of a specific metamodel or transformation may make the problem harder.
For instance, some endogenous transformations have a fixed point semantics and are called until a given type of element has been eliminated.
Each intermediate step produces elements of this type except the last one.
An example of such a transformation would eliminate for loops from a constraint program one nesting level at a time.

The objective of this paper is to provide a solution applicable with the current state of the art: actual domains and codomains cannot currently be 1) precisely computed, and 2) automatically checked.
Therefore, if an approximation (because of 1)) is computed it must be represented in a simple form that the user may understand quickly (because of 2): the user has to interpret it).
An example of such a simplification is the list of concepts (i.e., model element types coming from the metamodels) that are taken as input or produced as output of a transformation.
This is the first analysis that has been applied to the motivating example presented in Section~\ref{sec:motivatingExample} with relatively poor results if considered alone.

\subsection{Abstracting Rules}
\label{sec:abstractingRules}

Other kinds of information may be used to better understand what a transformation does.
ATL transformations are composed of rules that match source elements according to their type and some conditions (these form the source pattern of the rule), and that produce target elements of specific types (these form the target pattern of the rule).
A transformation analyzer may produce an abstract representation of a set of transformation rules.
This simplified description may take several forms.

One may think of representing the mapping between source and target metamodel concepts defined by the rules.
Model weaving may be used for this purpose as shown in \cite{BDA05,SAC06a}.
However, such a representation would be relatively verbose: there are as many mappings as rules, and the number of rules is typically close to the number of source or target concepts.

An additional simplification may be devised in the case of endogenous transformations in which elements are either copied (same target and source type) or mutated (different target and source types).
These actions may be applied on every element of a given type, or only under certain conditions.
Moreover, ATL lazy rules that are only applied if explicitly referenced (i.e., this is a kind of lazy evaluation) may also be used.
Table~\ref{tab:kindsOfRules} summarizes this classification of endogenous transformation rules.
The first dimension (in columns) is the kind of action (copy or mutation) that is performed by the rule.
The second dimension (in rows) corresponds to the cases in which the action is taken: always, under specific conditions, or lazily.
Corresponding examples of rules taken from the motivating example are given below.
No example of always or lazy mutation is given because there is no such case in the transformations of the motivating example.

\begin{table}
  \centering
  \caption{Classification of endogenous rules}
  \begin{tabular}{|c|c|c|}
  	\hline
  	& Copy & Mutation\\
  	\hline
  	Always &&\\
  	\hline
  	Conditionally &&\\
  	\hline
  	Lazily &&\\
  	\hline
  \end{tabular}
  \label{tab:kindsOfRules}
\end{table}

Listing~\ref{lst:alwaysCopy} gives a rule that always copies data types.
The target type (line 5) of such a rule is the same as its source type (line 3).
It is concept \emph{DataType} of the \emph{CPPivot} metamodel in this listing.
Moreover, it also copies all properties (e.g., source element name is copied to target element name at line 6).
However, property-level information is not always so simple to identify.
In many cases some properties are copied while others are recomputed.
In order to keep the information presented to the user simple, property-level information is ignored in the current implementation of the transformation analyzer.

\pagebreak	% TODO: this should not be a hardcoded page break because this is likely to cause problems if the paper is extended
\begin{lstlisting}[language=ATL, style=AMMA, numbers=left, caption={\emph{Always copy} rule example}, label=lst:alwaysCopy]
rule DataType {
	from
		s : CPPivot!DataType
	to
		t : CPPivot!DataType(
			name <- s.name
		)
}
\end{lstlisting}

A conditional copy happens when a copy rule has a filter or guard (i.e., a boolean expression that conditions the execution of the rule).
The rule of Listing~\ref{lst:condCopy} is similar to the rule presented above in Listing~\ref{lst:alwaysCopy} but has a guard specified at line 4.
This rule performs a conditional copy.

\begin{lstlisting}[language=ATL, style=AMMA, numbers=left, caption={\emph{Conditionally copy} rule example}, label=lst:condCopy]
rule SetDomain {
	from
		s : CPPivot!SetDomain (
			not s.parent.oclIsTypeOf(CPPivot!IndexVariable)
		)
	to
		t : CPPivot!SetDomain (
			values <- s.values
		)
}
\end{lstlisting}

Listing~\ref{lst:lazilyCopy} contains a lazy copy rule similar to the two previous rules of Listings~\ref{lst:alwaysCopy} and~\ref{lst:condCopy} but starting with keyword \emph{lazy} at line 1.
Additionally, the rule presented here extends another rule via rule inheritance.
This information is currently ignored during the abstraction process.

\begin{lstlisting}[language=ATL, style=AMMA, numbers=left, caption={\emph{Lazily copy} rule example}, label=lst:lazilyCopy]
lazy rule lazyBoolVal extends lazyExpression {
	from
		b : CPPivot!BoolVal
	to
		t : CPPivot!BoolVal(
			value <- b.value
		)
}
\end{lstlisting}

An example of conditional mutation is given in Listing~\ref{lst:condMutation}.
This rule is a mutation because the target type \emph{IntVal} (line 7) is different from the source type \emph{VariableExpr} (line 3).
It is conditional because there is a filter at line 4.

\begin{lstlisting}[language=ATL, style=AMMA, numbers=left, caption={\emph{Conditional mutation} rule example}, label=lst:condMutation]
rule VariableExpr2IntVal {
	from
		s : CPPivot!VariableExpr(
			s.declaration.oclIsTypeOf(CPPivot!EnumLiteral)
		)
	to
		t : CPPivot!IntVal(
			value <- s.declaration.getEnumPos
		)
}
\end{lstlisting}

\subsection{Implementing Transformation Analysis}

Transformation analysis is a case of Higher-Order Transformation \cite{ECMDA2009} (HOT): it is a transformation that takes as input another transformation to be analyzed, and produces as output a model containing the analysis result.
This HOT uses OCL expressions over the ATL metamodel, which is the metamodel of the language in which the transformations to analyze are written.
These expressions recognize the patterns presented in Section~\ref{sec:abstractingRules}.
Then, an analysis model is created that relates concepts of the pivot metamodel to recognized patterns.

\begin{figure}[htpb]
	\centering
	\includegraphics[bb=0 0 281 138]{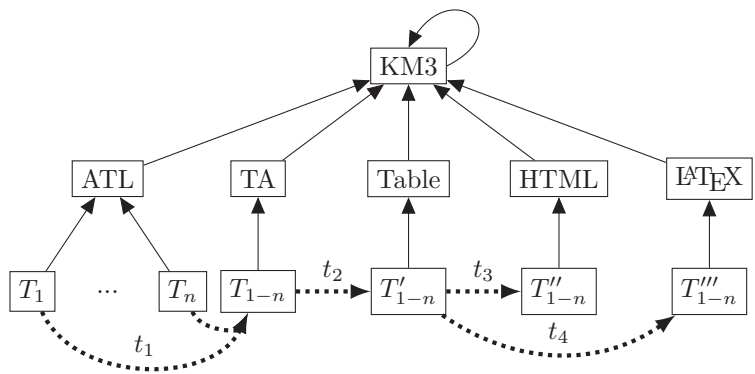}
	\caption{Transformation analysis and results rendering}
	\label{fig:rendering}
\end{figure}

The main objective is to deliver a result that a user may understand and interpret.
Consequently, special care was given to the rendering of the results.
Figure~\ref{fig:rendering} shows how the whole process is implemented.
It starts from a collection of $n$ ATL transformations $T_1$ to $T_n$ conforming to the ATL metamodel.
Transformation $t_1$ is applied to these transformations in order to obtain model $T_{1-n}$ conforming to the TA (for Transformation Analysis) metamodel.
This model contains the raw results of the analysis.

Then, transformation $t_2$ is applied in order to obtain model $T_{1-n}'$ that conforms to a generic Table metamodel.
This model may then be rendered to concrete display surfaces like HTML using transformation $t_3$, or \LaTeX  using transformation $t_4$.
The HTML rendering leverages the metamodels and transformation presented in \cite{MODELS2006MSMW}, and available from Eclipse.org.
The \LaTeX  rendering was specifically developed for the work presented in this paper.
The tables given as example in Section~\ref{sec:experiments} below have been generated automatically using the process depicted here.
All metamodels conform to the KM3 \cite{FMOODS06} (Kernel MetaMetaModel) metametamodel.

Although other techniques could have been used for the implementation, the whole transformation analysis and rendering process is defined in terms of models, metamodels, and transformations.
This is an example of the unification power of models \cite{journals/sosym/Bezivin05}.

%===============================================================================
\section{Experiments}
\label{sec:experiments}

\subsection{Application to the Motivating Example}
In the motivating example presented in
Section~\ref{sec:motivatingExample} (see Figure~\ref{fig:CP-Pivot}), we
consider five endogenous transformations achieving the following
reformulation tasks:
\begin{itemize}
  \item {\bf Class and objects removal.} This complex endogenous transformation is 
decomposed in two steps. The first step removes classes and does not copy their 
features. Variables with a class type are mutated in an untyped record
definition that is a duplication of the class features. Other variables --- with
a primitive type like integer, real or boolean --- are simply copied like other
elements not being contained in a class declaration. The second step
flattens record elements to get only variables with a primitive type.
  \item {\bf Enumeration removal.} Some CP solvers do not accept symbolic domains.
Thus, variables with a type being an enumeration are replaced by integer
variables with a domain ranging from 1 to the possible number of symbolic
values.
  \item {\bf Useless If removal.} Boolean expressions used as tests in conditional if
statements can be constant. In this case, it can be simplified, by removing
conditional if elements and keeping only the relevant collection of statements.
  \item {\bf For loops removal.} This reformulation task is implemented as a
fixed point transformation followed by the useless if removal transformation. In
the fixed point, each step removes only the deepest loops, i.e. loops that do
not contain other loops. To ease the loops removing task, this composite element
is replaced by another composite one being a conditional statement with an always
true boolean test (i.e., a block).
\end{itemize}

We have applied on this example the HOT presented in the previous
section. The results are detailed in the two following tables, which were 
automatically generated.

First, Table~\ref{tab:experimentalResultsIgn} presents the names of
ignored in and out concrete concepts for each analysed transformation.
These concepts are defined as concrete in the pivot metamodel, but they do
not appear in any OCL expression of transformations. We can see, there is only
one in ignored concept considering the record removal step. Indeed, this transformation
was written with the assumption of being launched after the class instantiation
transformation. Looking at the generated models, several concepts are missing,
such as Class and Record for the record removal transformation.

Second, Table~\ref{tab:experimentalResultsRef} gives more details on
what endogenous transformations really do. Each line corresponds to an endogenous
transformation analysis. Each column details the characteritics --- always,
conditionally and lazily --- of none, one, several, or all other
concepts. These characteristics are detailed for copy and mutation rules.

\newcommand{\largeColumnWidth}{4 cm}
\newcommand{\narrowColumnWidth}{1.5 cm}
\begin{table}
  \centering
  \caption{Experimental results: ignored elements}
	\begin{tabular}{|c|c|c|}
  \hline
  Transformation&Ignored in metaelements&Ignored out metaelements\\
  \hline
  classInstantiation& &Class\\
  \hline
  enumRemoval& &\parbox{\largeColumnWidth}{\centering EnumLiteral, Enumeration}\\
  \hline
  forallRemoval& & \\
  \hline
  recordRemoval&Class&Class, Record\\
  \hline
  uselessIfRemoval& & \\
  \hline
\end{tabular}

  \label{tab:experimentalResultsIgn}
\end{table}

\renewcommand{\largeColumnWidth}{2 cm}
\renewcommand{\narrowColumnWidth}{1.5 cm}
\begin{table}
  \centering
  \caption{Experimental results: referenced elements}
  \begin{sideways}
    \centering
		\begin{tabular}{|c|c|c|c|c|c|c|c|c|c|}
  \hline
  \textbf{Copy}&\parbox{\narrowColumnWidth}{\centering \textbf{lazily, cond.}}&\textbf{never}&\parbox{\narrowColumnWidth}{\centering \textbf{lazily, cond.}}&\textbf{cond.}&\textbf{never}&\textbf{always}&\textbf{cond.}\\
  \hline
  \textbf{Mutation}&\textbf{cond.}&\textbf{never}&\textbf{never}&\textbf{cond.}&\textbf{cond.}&\textbf{never}&\textbf{never}\\
  \hline
  \hline
 \textbf{classInstantiation}&NONE&Class&NONE&Variable&NONE&\parbox{\largeColumnWidth}{\centering EnumLiteral, Predicate, Enumeration, DataType, Model}&ALL OTHER\\
  \hline
  \textbf{enumRemoval}&NONE&\parbox{\largeColumnWidth}{\centering EnumLiteral, Enumeration}&NONE&Variable, VariableExpr&NONE&ALL OTHER&NONE\\
  \hline
  \textbf{forallRemoval}&Forall, VariableExpr&NONE &ALL OTHER&IndexVariable&NONE&\parbox{\largeColumnWidth}{\centering EnumLiteral, Predicate, Enumeration, Constant, DataType, Variable, Record, Class, Model, Array}&\parbox{\largeColumnWidth}{\centering SetDomain, IntervalDomain}\\
  \hline
  \textbf{recordRemoval}&NONE&Record&NONE&NONE&PropertyExpr&\parbox{\largeColumnWidth}{\centering EnumLiteral, Predicate, IndexVariable, Constraint, Enumeration, Constant, DataType, If, Model, Forall}&ALL OTHER\\
  \hline
  \textbf{uselessIfRemoval}&NONE&NONE&NONE&NONE&NONE&ALL OTHER&If\\
  \hline
\end{tabular}

  \end{sideways}
  \label{tab:experimentalResultsRef}
\end{table}

\subsection{Interpreting the Results}
\subsubsection{Typing source and target models.}
The results given by Table~\ref{tab:experimentalResultsIgn} can be
used to finely type authorized source and target models of the
transformation. The set of authorized element types can be obtained by
computing the difference between the set of all metamodel concepts and
those presented in Table~\ref{tab:experimentalResultsIgn}. It must be
noted that looking only at the concepts in source patterns is not enough,
since OCL navigation expressions can be used to explore and grab the
elements contained in one being removed.
Moreover, this information is only an approximation of the actual domain and codomain of the transformations, as described in Section~\ref{sec:transformationAnalysis}.

\subsubsection{Inferring partial transformation meaning.}
Considering Table~\ref{tab:experimentalResultsRef}, we can try to interprete the
discovered knowledge to infer the transformation meaning. In the case
of the class instantiation transformation, we can see that the only
concept never copied and never mutated is the class concept. Since it
is not in Table~\ref{tab:experimentalResultsIgn}, it appears within
OCL expressions, but it never appears within source patterns. It seems 
logical, since the aim of this transformation is to remove class
statements by expanding their features.
Then, variable elements are conditionally copied and conditionally mutated.
Indeed, variable types are checked to know if they must be copied (i.e., their type
is a primitive type) or if they must be mutated into record elements. Several concepts
are always copied and never mutated. They correspond to type definitions or the root
model concept, i.e. all concepts that can not be contained in a class. Finally
all other concepts are conditionally copied and never mutated. It is
checked they do not appear in a class before copying them.

Considering this knowledge, we can deduce that this transformation eliminates
class elements, even if they are used within OCL navigation
expressions. Variables are copied or mutated, whereas other elements
are copied (some of them under a condition). So, this transformation
mainly act on two types of elements: class and variable. We may use
the set of element types occurring in the target patterns to know the
sub-metamodel to which generated models conform.

Looking at the useless if removal transformation, we can easily infer its meaning.
Indeed, only the if statements are conditionnally copied, while all other elements
are always copied. Then, only some if statements are processed and might be removed.

\subsubsection{Discovering fixed point transformations.}

A transformation having a fixed point semantics may have its codomain
equal to its
domain. It may focus only on a few concepts to conditionally mutate and to
conditionally copy. All other concepts may be only copied. This pattern may
allow us to detect whether an endogenous transformation could be applied in
a fixed point scheme. In Tables~\ref{tab:experimentalResultsIgn}
and~\ref{tab:experimentalResultsRef} we see that the forall removal transformation
matches this pattern. Looking only at Table~\ref{tab:experimentalResultsRef},
we may think that the enumeration removal transformation is also a fixed point
transformation processing variables. However, Table~\ref{tab:experimentalResultsIgn}
shows that its main goal is to remove enumerations, because its domain and its
codomain are not equal (i.e., it removes all enumerations in one step).

%===============================================================================
\section{Discussions}
\label{sec:discussions}

\subsection{Application to Exogenous Transformations}
The approach presented in this paper could be extended to support exogenous
transformations. Thus, looking at the source patterns and all OCL expressions,
we can define the refined type of source models of a transformation (i.e., a more precise definition of its domain). To get the
refined type of target models (i.e., a more precise definition of the codomain), we just have to collect the set of concepts
occurring in target patterns.

Moreover, we can consider most endogenous transformations as exogenous transformations
between two sub-metamodels of the same metamodel. Then, the chaining of endogenous
transformations can be transformed into a problem of chaining exogenous transformations.
Inferring the meaning of an endogenous transformation may not be necessary (in most 
cases), since its main task may be to remove or add elements of a given type. However,
more complex endogenous transformations may be more difficult to finely chained, since
their meaning is necessary to understand how to use them. The knowledge collected in
Table~\ref{tab:experimentalResultsRef} is an attempt at achieving this goal with high-level
characteristics on concepts. However, this knowledge does not focus on how matchings are
performed in rules. Using a more detailed analysis, we could generate weaving models relating
to model transformations and then analyze them. However, these models would be more verbose than Table~\ref{tab:experimentalResultsRef}. We could also try to analyse OCL
expressions and mappings in transformation rules. Although, the cost and the difficulty of
our approach is almost negligible when compared to these deeper analysis.

\subsection{Debugging Transformations}
The knowledge discovered through our analysis transformation can be used in
debugging model transformations (exogenous or endogenous). Indeed, when a
metamodel contains many concepts, a software engineer may forget to define all
the corresponding rules. Thus the results from Table~\ref{tab:experimentalResultsIgn}
can be directly used, but also the column of
Table~\ref{tab:experimentalResultsRef} that corresponds to elements never copied
and never mutated. Other columns may also be useful to check that concepts are well
classified and no copy or mutation rule are missing.

The data in Table~\ref{tab:experimentalResultsRef} can also be used to discover
mistakes in naming metamodel concepts in some rules or helpers. Indeed, some
concepts of a metamodel may rarely have instances in models, and rules dealing
with them may not be called. Thus, no error occurs even if the transformation
contains some careless mistakes. In the case of our motivating example, we discovered
several ill-written rules and helpers dealing with specific CP concepts that do
not occur in our CP models.

%===============================================================================
\section{Conclusion}
\label{sec:conclusion}

In this paper, we addressed the problem of chaining model transformations. This
problem is illustrated on a pivot metamodel for Constraint Programming (CP) that is used for translations between
CP languages. Several
issues are tackled in order to safely chain transformations. Thus, a higher-order
transformation is proposed to statically analyze model transformations. It focuses
on source and target concepts, thus defining refined metamodels to which models
conform (i.e., more precise definitions of domains and codomains of model transformations). It also extracts some knowledge on how source concepts are processed and
assigns characteristics to each concept: always copied,
conditionally copied, lazily copied, never copied, always mutated, etc. Considering
these characteristics, we are able to find element types that are mainly processed.
This process is not accurate enough to exactly infer the meaning of model transformations (it is an abstraction),
but it allows us to assert some constraints on how to chain several endogenous transformations.
The contributions of this paper are of a different nature and complementary to the results presented in~\cite{Vignaga2008}.
That paper focuses on a type system for transformation chains, and considers that declared types are good enough, whereas in this paper we have investigated the problem of imprecise transformation typing.

A possible extension of the work presented in this paper would be to go beyond the discovery of hidden chaining constraints and to fully automatize transformation chaining. This automation process could be performed using Artificial Intelligence techniques. An optimization
problem can be defined to transform models from a source metamodel to another. The problem naturally
comes to find a path in a graph corresponding to a model of the transformations and their types. Some heuristics can be defined
to choose the best paths, which may contain as few redundant and as few useless steps as possible.

\bibliographystyle{plain}
\bibliography{cj_models_2009}

\begin{thebibliography}{10}

\bibitem{ATLLibrary}
{\em ATLAS Transformation Language (ATL) Library}.
\newblock \url{http://www.eclipse.org/m2m/atl/atlTransformations/}, 2009.

\bibitem{EclipseBook2007}
Krzysztof~R. Apt and Mark Wallace.
\newblock {\em {Constraint Logic Programming using Eclipse}}.
\newblock Cambridge University Press, New York, NY, USA, 2007.

\bibitem{journals/sosym/Bezivin05}
Jean B\'ezivin.
\newblock On the unification power of models.
\newblock {\em Software and System Modeling}, 4(2):171--188, 2005.

\bibitem{ChenouardPPDP2008}
Rapha\"el Chenouard, Laurent Granvilliers, and Ricardo Soto.
\newblock {Model-Driven Constraint Programming}.
\newblock In {\em Proceedings of ACM SIGPLAN PPDP}, pages 236--246, Valencia,
  Spain, 2008. ACM Press.

\bibitem{ChenouardSara2009}
Rapha\"el Chenouard, Laurent Granvilliers, and Ricardo Soto.
\newblock {Rewriting Constraint Models with Metamodels}.
\newblock In {\em Proceedings of SARA2009}. AAAI Press, 2009.

\bibitem{BDA05}
Marcos {Didonet Del Fabro}, Jean B\'ezivin, Fr\'ed\'eric Jouault, and Patrick
  Valduriez.
\newblock Applying generic model management to data mapping.
\newblock In {\em Proceedings of the Journ\'ees Bases de Donn\'ees Avanc\'ees
  (BDA05)}, 2005.

\bibitem{FrischIJCAI2007}
Alan~M. Frisch, Matthew Grum, Chris Jefferson, Bernadette {Mart\'inez
  Hern\'andez}, and Ian Miguel.
\newblock {The Design of ESSENCE: A Constraint Language for Specifying
  Combinatorial Problems}.
\newblock In {\em Proceedings of IJCAI}, pages 80--87, 2007.

\bibitem{OPL1999}
Pascal~Van Hentenryck.
\newblock {\em {The OPL Optimization Programming Language}}.
\newblock The MIT Press, 1999.

\bibitem{SCP07c}
Fr\'ed\'eric Jouault, Freddy Allilaire, Jean B\'ezivin, and Ivan Kurtev.
\newblock Atl: a model transformation tool.
\newblock {\em Science of Computer Programming}, 72(3, Special Issue on Second
  issue of experimental software and toolkits (EST)):31--39, 2008.

\bibitem{FMOODS06}
Fr\'ed\'eric Jouault and Jean B\'ezivin.
\newblock Km3: a dsl for metamodel specification.
\newblock In {\em Proceedings of 8th IFIP International Conference on Formal
  Methods for Open Object-Based Distributed Systems, LNCS 4037}, pages
  171--185, Bologna, Italy, 2006.

\bibitem{SAC06a}
Fr\'ed\'eric Jouault and Ivan Kurtev.
\newblock On the architectural alignment of atl and qvt.
\newblock In {\em Proceedings of the 2006 ACM Symposium on Applied Computing
  (SAC 06)}, pages 1188--1195, Dijon, France, 2006. ACM Press.

\bibitem{MTIP05b}
Fr\'ed\'eric Jouault and Ivan Kurtev.
\newblock Transforming models with atl.
\newblock In Jean-Michel Bruel, editor, {\em Satellite Events at the MoDELS
  2005 Conference: MoDELS 2005 International Workshops OCLWS, MoDeVA, MARTES,
  AOM, MTiP, WiSME, MODAUI, NfC, MDD, WUsCAM, Montego Bay, Jamaica, October
  2-7, 2005, Revised Selected Papers, LNCS 3844}, pages 128--138. Springer
  Berlin / Heidelberg, 2006.

\bibitem{Zinc2008}
Kim Marriott, Nicholas Nethercote, Reza Rafeh, Peter~J. Stuckey, Maria~Garcia
  de~la Banda, and Mark Wallace.
\newblock {The Design of the Zinc Modelling Language}.
\newblock {\em Constraints}, 13(3):229--267, 2008.

\bibitem{PugetSCIS1994}
Jean-Fran\c{c}ois Puget.
\newblock {A C++ Implementation of CLP}.
\newblock In {\em Proceedings of SPICIS'94}, Singapore, 1994.

\bibitem{PugetCP04}
Jean-Fran\c{c}ois Puget.
\newblock {Constraint Programming Next Challenge: Simplicity of Use}.
\newblock In {\em Proceedings of CP}, LNCS 3258, pages 5--8, 2004.

\bibitem{RossiElsevier2006}
Francesca Rossi, Peter {van Beek}, and Toby Walsh.
\newblock {\em {Handbook of Constraint Programming (Foundations of Artificial
  Intelligence)}}.
\newblock Elsevier Science Inc., New York, NY, USA, 2006.

\bibitem{SchulteGecode2006}
Christian Schulte and Guido Tack.
\newblock {Views and Iterators for Generic Constraint Implementations}.
\newblock In {\em Recent Advances in Constraints}, LNCS 3978, pages 118--132,
  2006.

\bibitem{SotoICTAI2007}
Ricardo Soto and Laurent Granvilliers.
\newblock {The Design of COMMA: An Extensible Framework for Mapping Constrained
  Objects to Native Solver Models}.
\newblock In {\em Proceedings of ICTAI}, pages 243--250. IEEE Computer Society,
  2007.

\bibitem{ECMDA2009}
Massimo Tisi, Fr\'ed\'eric Jouault, Piero Fraternali, Stefano Ceri, and Jean
  B\'ezivin.
\newblock On the use of higher-order model transformations.
\newblock In {\em Proceedings of the Fifth European Conference on Model-Driven
  Architecture Foundations and Applications (ECMDA)}, pages 18--33, 2009.

\bibitem{MODELS2006MSMW}
Eric V\'epa, Jean B\'ezivin, Hugo Bruneli\`ere, and Fr\'ed\'eric Jouault.
\newblock Measuring model repositories.
\newblock In {\em Proceedings of the Model Size Metrics Workshop at the
  MoDELS/UML 2006 conference, Genoava, Italy}, 2006.

\bibitem{Vignaga2008}
Andr\'es Vignaga, Fr\'ed\'eric Jouault, Mar\'ia~Cecilia Bastarrica, and Hugo
  Bruneli\`ere.
\newblock {Typing in Model Management}.
\newblock In {\em Proceedings of ICMT2009}, pages 197--212, Zurich,
  Switzerland, 2009.

\end{thebibliography}

\end{document}